\documentclass[11pt,a4paper]{article}
\usepackage[hyperref]{acl2019}
\usepackage{times}
\usepackage{latexsym}
\usepackage{makecell}
\usepackage{booktabs}
\usepackage{graphicx}
\usepackage{xcolor}

\usepackage{url}

\aclfinalcopy 
\def\aclpaperid{901} 

\title{Delta Embedding Learning}

\author{Xiao Zhang\footnotemark[1]\quad Ji Wu\footnotemark[1]\quad Dejing Dou\footnotemark[2]\\
		\footnotemark[1]\, Department of Electronic Engineering, Tsinghua University\\
		\footnotemark[2]\, Department of Computer and Information Science, University of Oregon\\
		\texttt{xiphzx@gmail.com}\\
		\texttt{wuji\_ee@mail.tsinghua.edu.cn}\\
		\texttt{dou@cs.uoregon.edu}}

\date{}

\begin{document}
\maketitle
\begin{abstract}
  Unsupervised word embeddings have become a popular approach of word representation in NLP tasks. However there are limitations to the semantics represented by unsupervised embeddings, and inadequate fine-tuning of embeddings can lead to suboptimal performance.
 We propose a novel learning technique called \textit{Delta Embedding Learning}, which can be applied to general NLP tasks to improve performance by optimized tuning of the word embeddings. A structured regularization is applied to the embeddings to ensure they are tuned in an incremental way. As a result, the tuned word embeddings become better word representations by absorbing semantic information from supervision without ``forgetting.'' We apply the method to various NLP tasks and see a consistent improvement in performance. Evaluation also confirms the tuned word embeddings have better semantic properties.
 
\end{abstract}

\section{Introduction} 

Unsupervised word embeddings have become the basis for word representation in NLP tasks. Models such as skip-gram \cite{mikolov2013efficient} and Glove \cite{pennington2014glove} capture the statistics of a large corpus and have good properties that corresponds to the semantics of words \cite{mikolov2013distributed}. However there are certain problems with unsupervised word embeddings, such as the difficulty in modeling some fine-grained word semantics. For example words in the same category but with different polarities are often confused because those words share common statistics in the corpus \cite{faruqui:2015:NAACL,mrkvsic2016counter}.

In supervised NLP tasks, these unsupervised word embeddings are often used in one of two ways: keeping fixed or using as initialization (fine-tuning). The decision is made based on the amount of available training data in order to avoid overfitting. Nonetheless, underfitting with keeping fixed and certain degrees of overfitting with fine-tuning is inevitable. Because this all or none optimization of the word embeddings lacks control over the learning process, the embeddings are not trained to an optimal point, which can result in suboptimal task performance,  as we will show later.

In this paper, we propose \textit{delta embedding learning}, a novel method that aims to address the above problems together: using regularization to find the optimal fine-tuning of word embeddings. Better task performance can be reached with properly optimized embeddings. At the same time, the regularized fine-tuning effectively combines semantics from supervised learning and unsupervised learning, which addresses some limitations in unsupervised embeddings and improves the quality of embeddings.

Unlike retrofitting \cite{yu2014improving,faruqui:2015:NAACL}, which learns directly from lexical resources, our method provides a way to learn word semantics from supervised NLP tasks. Embeddings usually become task-specific and lose its generality when trained along with a model to maximize a task objective. Some approach tried to learn reusable embeddings from NLP tasks include multi-task learning, where one predicts context words and external labels at the same time \cite{tang2014learning}, and specially designed gradient descent algorithms for fine-tuning \cite{yang2015supervised}. Our method learns reusable supervised embeddings by fine-tuning an unsupervised embeddings on a supervised task with a simple modification. The method also makes it possible to examine and interpret the learned semantics.

The rest of the paper is organized as follows. Section \ref{sec:method} introduces the \textit{delta embedding learning} method. Section \ref{sec:exp} applies the method to NLP tasks, and the learned embeddings are evaluated and analyzed in section \ref{sec:eval}.

\section{Methodology} 
\label{sec:method}

\begin{figure}
\centering
\includegraphics[keepaspectratio=true, scale=.4]{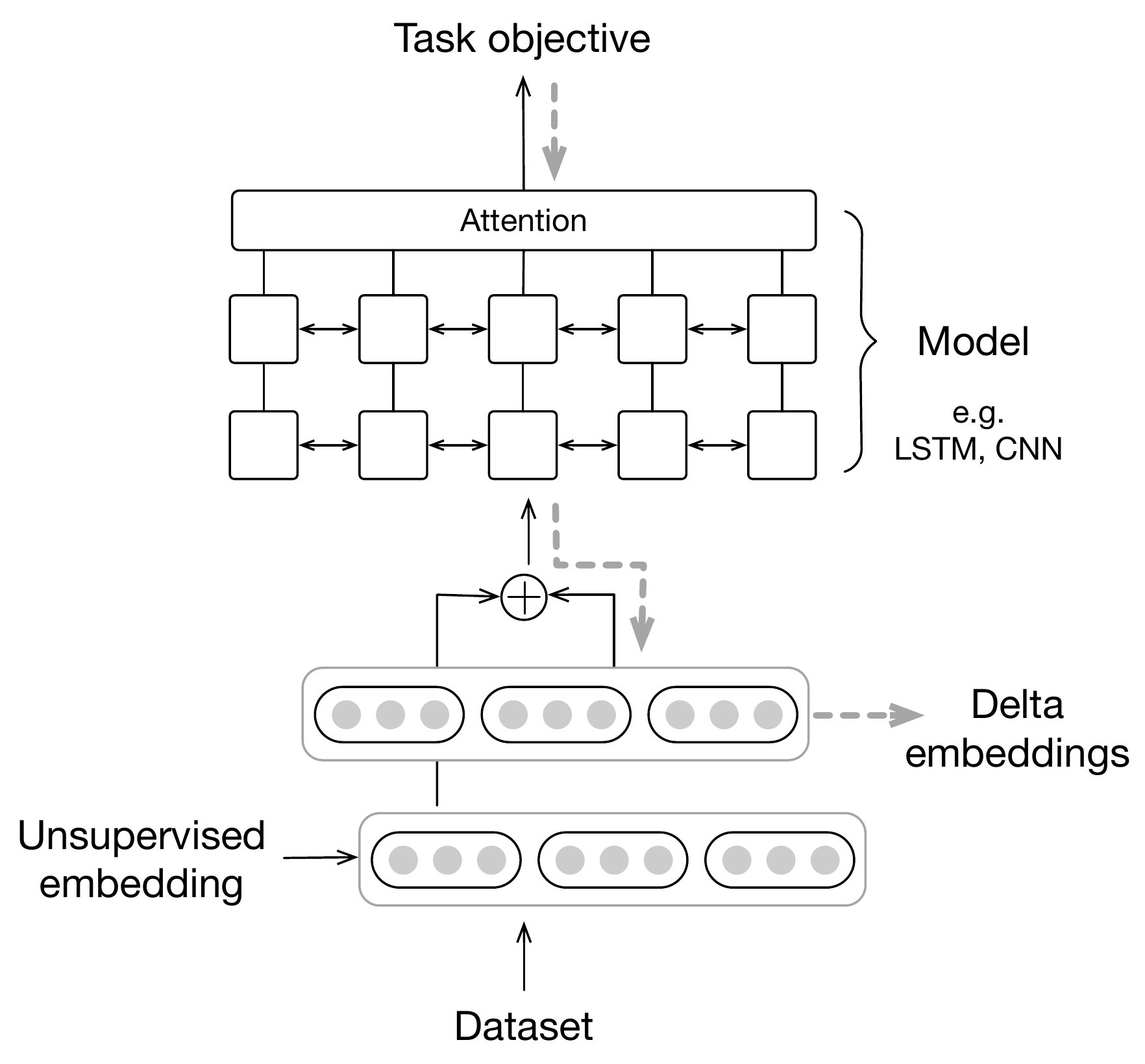}
    \caption{Delta embedding learning in a supervised NLP task. Solid line: forward model computation. Dashed line: learning of delta embeddings through back propagation}
    \label{fig:sys}
\end{figure}

\subsection{Delta embedding learning}

The aim of the method is to combine the benefits of unsupervised learning and supervised learning to learn better word embeddings. An unsupervised word embeddings like skip-gram, trained on a large corpus (like Wikipedia), gives good-quality word representations. We use such an embedding $\mathbf{w}_{unsup}$ as a starting point and learn a delta embedding $\mathbf{w}_{\Delta}$ on top of it:
\begin{equation}     
	\mathbf{w} = \mathbf{w}_{unsup} + \mathbf{w}_{\Delta}. 
\end{equation}

The unsupervised embedding $\mathbf{w}_{unsup}$ is fixed to preserve good properties of the embedding space and the word semantics learned from large corpus. Delta embedding $\mathbf{w}_{\Delta}$ is used to capture discriminative word semantics from supervised NLP tasks and is trained together with a model for the supervised task.  In order to learn only useful word semantics rather than task-specific peculiarities that results from fitting (or overfitting) a specific task, we impose $L_{21}$ loss, one kind of structured regularization on $\mathbf{w}_{\Delta}$:
\begin{equation}
\label{equ:loss}
	loss = loss_{task} + c\sum_{i=1}^n(\sum_{j=1}^dw_{\Delta ij}^2)^{\frac{1}{2}} 
\end{equation}

The regularization loss is added as an extra term to the loss of the supervised task.

The effect of $L_{21}$ loss on $\mathbf{w}_{\Delta}$ has a straightforward interpretation: to minimize the total moving distance of word vectors in embedding space while reaching optimal task performance. The $L_2$ part of the regularization keeps the change of word vectors small, so that it does not lose its original semantics. The $L_1$ part of the regularization induces sparsity on delta embeddings, that only a small number of words get non-zero delta embeddings, while the majority of words are kept intact. The combined effect is selective fine-tuning with moderation: delta embedding captures only significant word semantics that is contained in the training data of the task while absent in the unsupervised embedding.

\subsection{Task formulation}
Delta embedding learning is a general method that theoretically can be applied to any tasks or models that use embeddings. Figure \ref{fig:sys} is an illustration of how the method is applied. The combined delta embedding and unsupervised embedding is provided to a model as input. The delta embedding is updated with the model while optimizing the loss function in (\ref{equ:loss}). The model is trained to maximize task performance, and the produced delta embedding when combined with the unsupervised embedding becomes an improved word representation in its own right.

\section{Experiments on NLP tasks}	
\label{sec:exp}
\begin{table*}[t]
\fontsize{10}{12} \selectfont
\begin{center}
\begin{tabular}{lcccccccc}
	\hline \makecell[l]{Regularization \\ coefficient} & rt-polarity & \makecell{KMR} & \multicolumn{2}{c}{\makecell{SQuAD \\ EM\quad\quad F1}} & \multicolumn{3}{c}{\makecell{MultiNLI \\ Genre\quad M\quad Mis-M}} & SNLI\\ \hline

$0$ (finetune)     & 78.61 & 68.43 & 64.29 & 74.35 & 69.3 & 61.2 & 62.1 & 57.2\\
$\infty$ (fixed)& 76.66 & 66.72 & 67.94 & 77.33 & 69.5 & 62.6 & 64.0 & 59.7\\
$10^{-3}$               & 76.17 & 67.01 & 68.08 & 77.56 & 69.5 & 63.0 & 63.6 & 60.2\\
$10^{-4}$               & \bf{79.30} & 67.97 & \bf{68.45} & \bf{78.12} & \bf{71.5}& 63.4& \bf{64.3} & \bf{60.6}\\
$10^{-5}$               & 78.71 & \bf{68.96} & 66.48 & 76.31 & 70.9 & \bf{63.6} & 63.8 & 59.7\\
\hline
\end{tabular}
\end{center}
\caption{\label{perf-table} Performance of different embedding training methods on various NLP tasks. Numbers represent model accuracy (in percentage) on each task , except for SQuAD}
\end{table*}

\label{experiments}

We conduct experiments on several different NLP tasks to illustrate the effect of delta embedding learning on task performance. 

\subsection{Experimental setup}

\paragraph{Sentiment analysis} We performed experiments on two sentiment analysis datasets: rt-polarity (binary) \cite{pang2005seeing} and Kaggle movie review (KMR, 5 class) \cite{socher2013recursive}. For rt-polarity, we used a CNN model as in \cite{kim2014convolutional}. For KMR an LSTM-based model is used.

\paragraph{Reading comprehension} We used the Stanford Question Answering Dataset (SQuAD, v1.1) \cite{2016arXiv160605250R} and the Bi-directional Attention Flow (BiDAF) \cite{seo2016bidirectional} model. The original hyperparameters are used, except that character-level embedding is turned off to help clearly illustrate the effect of word embeddings.

\paragraph{Language inference} The MultiNLI \cite{N18-1101} and SNLI \cite{snli:emnlp2015} datasets are used for evaluation of the natural language inference task. We use the ESIM model, a strong baseline in \cite{N18-1101}. As MultiNLI is a large dataset, we use a subset (``fiction" genre) for training to simulate a moderate data setting, and use development set and SNLI for testing.

\paragraph{Common setup} For all the experiments, we used Glove embeddings pre-trained on Wikipedia and Gigaword corpus\footnote{https://nlp.stanford.edu/projects/glove/} as they are publicly available and frequently used in NLP literature. Dimensions of word embeddings in all models are set to 100.

\subsection{Results}

The task performance of models with different embedding learning choices is reported in Table \ref{perf-table}. All  initialized with unsupervised pre-trained embeddings, comparison is made between fine-tuning, keeping fixed and tuning with delta embeddings. For delta embeddings, there is one hyperparameter $c$ that controls the strength of regularization. We empirically experiment in the range of $[10^{-5}, 10^{-3}]$.

In all the tasks delta embedding learning outperforms conventional methods of using embedding. As embeddings is the only variable, it shows delta embedding learning learns better quality embeddings that results in better task performance. 

Roughly two kinds of scenarios exist in these tasks. For easier tasks like sentiment analysis underfitting is obvious when keeping embeddings fixed. Harder tasks like reading comprehension on the other hand clearly suffer from overfitting. In both situations delta embeddings managed to balance between underfitting and overfitting with a more optimal tuning.  

For the hyper-parameter choice of regularization coefficient $c$, we found it fairly insensitive to tasks, with $c=10^{-4}$ achieving the best performance in most tasks. 

The results indicate that delta embedding learning does not require the decision to fix the embedding or not in an NLP task, as delta embedding learning always harvests the best from unsupervised embeddings and supervised fine-tuning, regardless of the amount of labeled data. 

\section{Embedding evaluation}
\label{sec:eval}

To validate the hypothesis that better performance is the result of better embeddings, we examine the properties of embeddings tuned with delta embedding learning. Word embedding from the BiDAF model is extracted after training on SQuAD, and is compared with the original Glove embedding. 

The motivation of investigating embeddings trained on SQuAD is because reading comprehension is a comprehensive language understanding task that involves a rather wide spectrum of word semantics. Training on SQuAD tunes a number of word embeddings which results in non-trivial changes of embedding properties on the whole vocabulary level, which we can validate with embedding evaluation tests. As for simpler tasks like sentiment analysis, we observe that they tune fewer words and the effects are less visible.

\subsection{QVEC}
QVEC \cite{tsvetkov2015evaluation} is a comprehensive evaluation of the quality of word embeddings by aligning with linguistic features. We calculated the QVEC score of learned embeddings (Table \ref{qvec-table}).

\begin{table}[h]
\small
\begin{center}
\begin{tabular}{lcc}	
\hline
Embedding & QVEC score & Relative gain\\
\hline
 Glove & 0.37536   & - \\
finetune & 0.37267 & $-2.7\cdot 10^{-3}$\\
delta@$10^{-3}$           & 0.37536 & $3.0\cdot 10^{-6}$\\
delta@$10^{-4}$           & \bf{0.37543} & $7.5\cdot 10^{-5}$\\
delta@$10^{-5}$           & 0.37332 & $-2.0\cdot 10^{-3}$\\
\hline
\end{tabular}
\end{center}
\caption{\label{qvec-table} QVEC scores of learned embeddings}
\end{table}

Using the original Glove embedding as reference, unconstrained finetune decreases the QVEC score, because the embedding overfits to the task, and some of the semantic information in the original embedding is lost. Delta embedding learning ($c=10^{-4}$) achieves the best task performance while also slightly increases the QVEC score. The change in score is somewhat marginal, but can be regarded as a sanity check: delta embedding learning does not lower the quality of the original embedding (in other words, it does not suffer from catastrophic forget). Also, as the QVEC score is strongly related to downstream task performance, it also means that delta-tuned embedding is no less general and universal than the original unsupervised embedding. 

\subsection{Word similarity}
Word similarity is a common approach for examining semantics captured by embeddings. We used the tool in \cite{faruqui-2014:SystemDemo} to evaluate on 13 word similarity datasets. 
\begin{table}[t]
\small
\begin{center}
\begin{tabular}{lcccc}
\hline Correlation & Glove & finetune & \makecell{delta\\@$10^{-4}$} & $\Delta$\\
\hline
WS-353       & 0.555 & 0.545 &  \bf{0.563}  & +                 \\
WS-353-SIM   & 0.657 & 0.659 &  \bf{0.667}  & +                 \\
WS-353-REL   & 0.495 & 0.485 &  \bf{0.506}  & +                 \\
MC-30        & 0.776 & 0.764 &  \bf{0.783}  & +                 \\
RG-65        & \bf{0.741} & 0.736  & 0.740  & -                  \\
Rare-Word    & 0.391 & 0.377 & \bf{0.392}   & +                 \\
MEN          & 0.702 & \bf{0.703} & \bf{0.703}   & +                 \\
MTurk-287    & 0.632 & 0.625 & \bf{0.635}   & +                 \\
MTurk-771    & 0.574 & \bf{0.577}  & 0.576  & +                 \\
YP-130       & 0.460 & \bf{0.475}  & 0.467  & +                 \\
SimLex-999   & 0.292 & \bf{0.304}  & 0.295  & +                 \\
Verb-143     & 0.302 & 0.305 & \bf{0.315}   & +                 \\
SimVerb-3500 & 0.169 & \bf{0.176} & 0.171   & +             \\
\hline
\end{tabular}
\end{center}
\caption{\label{eval-table} Evaluation of embedding by word pair similarity ranking.}
\end{table}

Showed in Table \ref{eval-table}, delta embedding trained with $c=10^{-4}$ has the best performance in over half of the benchmarks. When compared to the original Glove embedding, unconstrained fine-tuned embedding gets better at some datasets while worse at others, indicating that naive fine-tuning learns some semantic information from the task while ``forgetting" some others. Delta embedding learning however, achieves better performance than Glove embedding in all but one datasets (negligible decrease on RG-65, see the last column of Table \ref{eval-table}). This shows that delta embedding learning effectively learns new semantics from a supervised task and adds it to the original embedding in a non-destructive way. The quality of embedding is improved.

\subsection{Interpreting word semantics learning} 
The formulation of delta embeddings makes it possible to help analyze word semantics learned in a supervised task, 
    regardless of the model used.
\begin{table}
\small
\begin{tabular}{ll}	
\hline
\makecell[bl]{Sentiment\\ Analysis} & \makecell[l]{neither still unexpected nor bore \\
lacking worst suffers usual moving \\
works interesting tv fun smart}  \\[13pt]
\makecell[bl]{Reading\\ Comprehension}& \makecell[l]{why another what along called \\
whose call which also this if not \\
occupation whom but he because into} \\[13pt]
\makecell[bl]{Language\\ Inference} & \makecell[l]{not the even I nothing because that you \\
 it as anything only was if want forget \\
 well be so from does in certain could} \\

\hline	
\end{tabular}
\caption{\label{top-delta-norm} Words with the largest norm of delta embedding in different tasks}
\end{table}

To answer the question ``What is learned in the task?", the norm of delta embeddings can be used to identify which word has a significant newly learned component. In Table \ref{top-delta-norm}, for instance, words with a  sentiment like ``bore" and ``fun" are mostly learned in sentiment analysis tasks. In reading comprehension, question words like ``what" and ``why" are the first to be learned , after that are words helping to locate possible answers like ``called," ``another," and ``also." 
\begin{table}[h]
\small
\begin{tabular}{ll}	
\hline
 & Nearest neighbors of word ``\it{not}''\footnotemark \\[5pt]
\makecell[l]{Before \\ training} & \makecell[l]{(+) good always clearly definitely well able\\
(-) nothing yet none}\\[10pt]
\makecell[bl]{After\\ training} & \makecell[l]{(+) sure \\
(-) nothing yet none bad lack unable nobody \\
\quad less impossible unfortunately Not rarely} \\

\hline	
\end{tabular}
\caption{\label{shift-nn} The position shift of word ``not'' in embedding space}
\end{table}
\footnotetext{only showing words with a polarity}

The semantics learned in a word can be represented by its shift of position in the embedding space (which is the delta embedding). We found the semantics learned are often discriminative features. Use the word ``not'' as an example, after training it clearly gains a component representing negativity, and differentiates positive and negative words much better (Table \ref{shift-nn}). These discriminative semantics are sometimes absent or only weakly present in co-occurrence statistics, but play a crucial role in the understanding of text in NLP tasks. 

\section{Conclusion}   
We proposed delta embedding learning, a supervised embedding learning method that not only improves performance in NLP tasks, but also learns better universal word embeddings by letting the embedding ``grow" under supervision.

Because delta embedding learning is an incremental process, it is possible to learn from a sequence of tasks, essentially ``continuous learning" \cite{parisi2018continual} of word semantics. It is an interesting future work and will make learning word embeddings more like human learning a language.

\section*{Acknowledgments}
This research is partially supported by the National Key Research and Development Program of China (No.2018YFC0116800) and the NSF grant CNS-1747798 to the IUCRC Center for Big Learning. 

\bibliography{delta-emb}
\bibliographystyle{acl_natbib}

\end{document}